\documentclass{article}

\PassOptionsToPackage{numbers,compress}{natbib}
\usepackage[final]{neurips_2020}

\usepackage[utf8]{inputenc} % allow utf-8 input
\usepackage[colorlinks=true,citecolor=blue]{hyperref}       % hyperlinks
\usepackage{hyperref}       % hyperlinks
\usepackage{url}            % simple URL typesetting
\usepackage{booktabs}       % professional-quality tables
\usepackage{amsfonts}       % blackboard math symbols
\usepackage{nicefrac}       % compact symbols for 1/2, etc.
\usepackage{microtype}      % microtypography
\usepackage{graphicx}
\usepackage{amsmath}
\usepackage{placeins}
\usepackage{todonotes}

\newcommand{\forecast}{\mathcal{T}_f}

\newcommand{\train}{\mathcal{T}_t}
\newcommand{\counts}{c_i^t}
\newcommand{\baselinelow}{b_{i, \text{low}}^t}
\newcommand{\baselineupp}{b_{i, \text{upp}}^t}
\newcommand{\baseline}{b_i^t}
\newcommand{\ebpscore}{F_{\text{EBP}}}
\newcommand{\asymscore}{F_{\text{ASYM}}}
\newcommand{\aggbaselineupp}{B_{S, \text{upp}}}
\newcommand{\aggbaselinelow}{B_{S, \text{low}}}

\usepackage{wrapfig}
\usepackage{multirow}
\hypersetup{
    colorlinks=true,
    linkcolor=red,
    filecolor=red,      
    urlcolor=blue,
}

\usepackage[ruled,vlined]{algorithm2e}
\usepackage{cleveref}
\crefname{section}{§}{§§}
\Crefname{section}{§}{§§}
\crefformat{section}{§#2#1#3} % Removes spacing after \S
\usepackage{subcaption}
\usepackage[title]{appendix}

\title{An Expectation-Based Network Scan Statistic for a COVID-19 Early Warning System}

\usepackage{authblk}

\renewcommand*{\authorcr}{\protect\\}

\author[1,2]{Chance Haycock}
\author[1,2]{Edward Thorpe-Woods}
\author[1]{James Walsh}
\author[2]{Patrick O'Hara}
\author[1]{Oscar Giles}
\author[1,2]{Neil Dhir}
\author[1,2,3]{Theodoros Damoulas}

\affil[1]{The Alan Turing Institute \authorcr \{\tt chaycock, ethorpe-woods, jwalsh, ogiles, ndhir\}@turing.ac.uk\vspace{4pt}}
\affil[ ]{Departments of \textsuperscript{2}Computer Science and \textsuperscript{3}Statistics, University of Warwick \authorcr \{\tt patrick.h.o-hara, t.damoulas\}@warwick.ac.uk\vspace{4pt}}

\begin{document}

\maketitle
\vspace{-1em}
\begin{abstract}
One of the Greater London Authority's (GLA) response to the COVID-19 pandemic brings together multiple large-scale and heterogeneous datasets capturing mobility, transportation and traffic activity over the city of London to better understand `busyness' and enable targeted interventions and effective policy-making. As part of \href{https://www.turing.ac.uk/research/research-projects/project-odysseus-understanding-london-busyness-and-exiting-lockdown}{Project Odysseus} we describe an early-warning system and introduce an expectation-based scan statistic for networks to help the GLA and Transport for London, understand the extent to which populations are following government COVID-19 guidelines. We explicitly treat the case of geographically \emph{fixed} time-series data located on a (road) network and primarily focus on \emph{monitoring} the dynamics across large regions of the capital. Additionally, we also focus on the detection and reporting of significant spatio-temporal regions. Our approach is extending the \emph{Network Based Scan Statistic} (NBSS) by making it expectation-based (EBP) and by using stochastic processes for time-series forecasting, which enables us to quantify metric uncertainty in both the EBP and NBSS frameworks. We introduce a variant of the metric used in the EBP model which focuses on identifying space-time regions in which activity is \emph{quieter} than expected.
%\footnote{To avoid breaking the double-blind review we do not include additional details of the greater project in the manuscript or appendix but we will do so if accepted, including the code.}
\end{abstract}

\section{Introduction}
\label{sec:intro}

In response to the current COVID-19 pandemic, governments of half the world's population have enacted some form of ``lock-down" \citep{muhammad2020covid}; a non-pharmaceutical approach to suppressing the spread of the virus by restricting social interaction -- a primary source of COVID-19 transmission \citep{badr2020association, matrajt2020evaluating}. With some countries now beginning the relaxation of these measures, as well as a looming second-wave; the monitoring of daily human activity has never been more important for policy makers. As we transition from nationwide lock-downs (e.g. in the UK, France, Spain and Italy) and enter the recovery period it is of paramount importance to understand to what extent human activity is returning to normal, enabling targeted interventions and effective policy-making, to prevent the resurgence of COVID-19 and to aid the economic recovery. Here, we describe the implementation of an \emph{early warning system} that enables the \emph{monitoring} of human activity levels and ultimately, the \emph{detection} of regions where levels are higher than expected. This enables us to understand mobility and activity in near real-time; assist public health planners, epidemiologists, policy-makers and support and help to optimise the economic recovery. Monitoring the collective movement of millions of people has been done before, to e.g. quantify malaria transmission rates \citep{wesolowski2012quantifying}. But clearly, large-scale computational studies such as this, raise substantial ethical concerns. Alas, note all data used in this project is anonymised and the system undergoes continuous review by The Alan Turing Institute's Ethical Advisory Group.

\section{Background}
The field of scan statistics is vast and is used for detecting anomalous clusters in spatial or spatio-temporal data \citep{diggle2013statistical, daley2007introduction, gelfand2010handbook, baddeley2015spatial}. The spatial scan statistic was introduced by \citet{kulldorff1997spatial} and one of many extensions is the expectation-based version by \citet{neill2005detection, neill2019machine}. In the latter, simultaneous monitoring of a large number of spatially localised time-series \citep{neill2005detection} is undertaken, to detect emerging clusters (i.e. spatial patterns). By computing the expected count for each recent day, for each spatial location, and then comparing those to more recent counts, \citet{neill2005detection} improve timeliness, accuracy and the spatial resolution of detection. \citet{shiode2020network} recently introduced the network-based scan-statistic (NetScan), an extension of the original by \citet{kulldorff1997spatial}. NetScan is one of many statistical methods for analysing spatial patterns of points on a network of lines, such as road accident locations along a road network \citep{baddeley2020analysing}. But focusing on the former for brevity, NetScan departs from the original scan-approach by replacing a standard planar search window with a network-based search window \citep[\S 2.1]{shiode2020network}. This is then used to sweep across the entire extent of the street network in the area of investigation. By extending the EBP to the network space, see \cref{sec:method}, we leverage the considerable statistical power therein, to the often complex geometries found in networks -- see \cref{fig:westminster}.

%  -- ``a set of connected line segments that extends from a reference point on a street network with its total length ranging from zero to a pre-determined maximum length''

% \begin{wrapfigure}{r}{0.55\linewidth}
% \vspace{-10pt}
% \begin{minipage}{\linewidth}
% {\centering
%     \begin{algorithm}[H]
%         \SetAlgoLined
%         \KwResult{Collection of scored space-time regions}
%         Define collection of spatially localised time-series data
%         Forecast each series $W$ time steps ahead\;
%         Define space-time regions $\mathcal{S}$ \;
%         \For{region $S$ in $\mathcal{S}$}{
%             sum baseline counts in region $S$\;
%             sum actual counts in region $S$\;
%             compute metric\;
%         }    
%         \caption{General Scan Statistic Pipeline \nd{finish}}
%     \end{algorithm}
% \par
% }
% \end{minipage}
% \vspace{-10pt}
% \end{wrapfigure}

% Much of this previous work is devoted to finding \emph{the} spatial (-temporal) region with increased activity and not necessarily the monitoring of a larger region in space. \citet{diggle2013statistical} makes the link between this type of data and spatio-temporal point processes, namely the Log-Gaussian Cox process (LGCP).

% EXPLAIN LNK. These techniques provide a way to visualise the ..., but does not automatically detect significant regions. Extending the LGCP to the spatio-temporal domain is also non-trivial, so we opt for ....

\section{Method}
\label{sec:method}
Consider a time-series, sampled at every hour, at a set of $N_s$ fixed spatial locations $\mathcal{X} = \{\mathbf{x}_i\}_{i=1}^{N_s}$ s.t. $\mathbf{x}_i \in \mathbb{R}^{2}$ on a network $\mathcal{G} = (V, E)$ (see \cref{sec:spatial_locs}) spanning up to the present time $T$ (e.g. vehicle count data from $N_s$ fixed road sensors). Where $V$ is a set of vertices and $E$ is a set of edges. Our goal is to scan over the most recent $W$ hours worth of data and identify spatio-temporal clusters by comparing expected counts $\baseline \in \mathbb{Z}_+$ with true counts $\counts \in \mathbb{Z}_+ $. Our approach has three distinct chapters:  first (\cref{sec:forecasting}), we generate the baselines $\baseline$ -- expected count data which represents typical behaviour. This provides the null hypothesis data, to compare to actual count data to locate emerging regions of busyness; %. Baselines are founding use time-series forecasting methods. \citet{neill2009expectation} compares the suitability of a number of simplistic methods but performance is very much dependent on the scientific question being asked.
second (\cref{sec:grid}), a grid is defined on the search domain and hence the search regions (remember, our search is over a road-network $\mathcal{G}$); third (\cref{sec:cluster_metrics}), a metric is defined which compares the expected and true counts. Consequently we are able to assign scores to each search region and use these to either monitor the entire region or report back on significant increases/decreases in activity.
    
%The proposed framework can be divided into three distinct parts; firstly, we generate the baselines $\baseline$ - expected count data which represents typical behaviour. This provides the null hypothesis data, to compare to actual count data to locate emerging regions of busyness. baselines are produced using some method of time-series analysis that can accurately forecast $W$ time steps ahead; \citet{neill2009expectation} compares the suitability of a number of simplistic methods but performance is very much dependent on the scientific question being asked. With baselines in place, the second step is to define the grid on the search domain and hence the search regions. When searching over a plane, there are many ways to do this \todo{such as?}. Thirdly, by defining a metric which compares the expected and true counts, we are able to assign scores to each search region and use these to either monitor the entire region or report back on significant increases/decreases in human activity.

\subsection{Forecasting}
\label{sec:forecasting}

We first generalise the forecasting by allowing disjoint intervals of time used for training and forecasting periods; these are denoted by time-indexing sets $\train$ and $\forecast$ respectively. After accounting for day-of-the-week variation, we then use true counts $\{\counts \ | \ t \in \train\}$ to learn baseline estimates $\{ \baseline \ | \ t \in \forecast \}$. In the context of a COVID-19 monitoring system, specifying $\train$ is dependent on the subtle scientific question that is being asked. For example, one could specify $\train$ as a time period of \emph{mid-lockdown} data to identify regions where traffic levels are higher than expected. %In the context of ``emerging cluster detection", most methods in the literature use contiguous training and forecast periods; we opt against this to avoid possible growing trends in the training data skewing the predicted counts.

The second generalisation incorporates uncertainty into the forecast through the use of Gaussian processes (GP) \citep{GPBook}. The GP framework yields predictions $\baseline$ that are Gaussian distributed with mean $\mu_i^t$ and standard deviation $\sigma_i^t$ \citep{GPBook} (when treating the counts as continuous variables). Consequently we estimate an upper and lower bound for each $\baseline$ by adding/subtracting some pre-determined number of $\sigma_i^t$ from the mean $\baseline = \mu_i^t$. We denote these upper and lower bounds by $\baselineupp$, $\baselinelow$ respectively and ultimately scan over search regions with data $\{ (\counts, \baseline, \baselineupp, \baselinelow) \ | \ t \in \forecast,\ \mathbf{x}_i \in \mathcal{X} \}$. Further, most human activity, such as traffic flow, is expected to exhibit both daily and weekly periodic variation which suggests the use of a tailored kernel of the form:
\begin{equation}
    k(x, y) = k_{\text{periodic}}^{\text{daily}}(x, y) \times k_{\text{periodic}}^{\text{weekly}}(x, y) + k_{\text{RBF}}(x, y) + k_{\text{white}}(x, y).
\end{equation}
We explicitly initialise the periods of the periodic components to 24h and 168h respectively. In addition, we compare the GP approach to the very successful Holt-Winters method used by \citet{neill2009expectation} (though they find that performance is very much dependent on the scientific question being asked). See \cref{fig:jam_cam_forecast}, \cref{sec:appendix_forecasting}, for an example forecast.

\newpage
\subsection{Grid definitions}
\label{sec:grid}
% \nd{This section needs to be much shorter.}
 %For the planar scan, we generate our collection of search regions by dividing the domain $\mathcal{D}\in\mathbb{R}^2$ into $N^2$ equally sized rectangular regions; typically $N = 2^j$ for some small $j\in\mathbb{N}_+$. This will depend on the domain of interest and the required output resolution.
 %We scan over \emph{all} rectangles that cover up to half of the domain on this grid. Whilst this removes problem of choosing a suitable partition size, the number of spatial regions scales with $\mathcal{O}(N^4)$. %Taking into the account the temporal nature of the scan, the number of spatio-temporal search regions is then $\mathcal{O}(|\forecast |N^4)$. Faster versions of the scan have been proposed by \citet{Multi-Dimensional-General-stat} but speed-ups are made by avoiding searching all regions, making it impossible to \emph{monitor} the entire spatial domain.
 %However, when using network-based data sources, most of these search regions are either empty or sparsely populated with sensor readings. It is this observation that motivates the use of a network-based scan statistic.
     
\begin{wrapfigure}[15]{r}{0.4\textwidth}
    \vspace{-1em}
    \centering
    % \centercaption
    \includegraphics[width=0.4\textwidth]{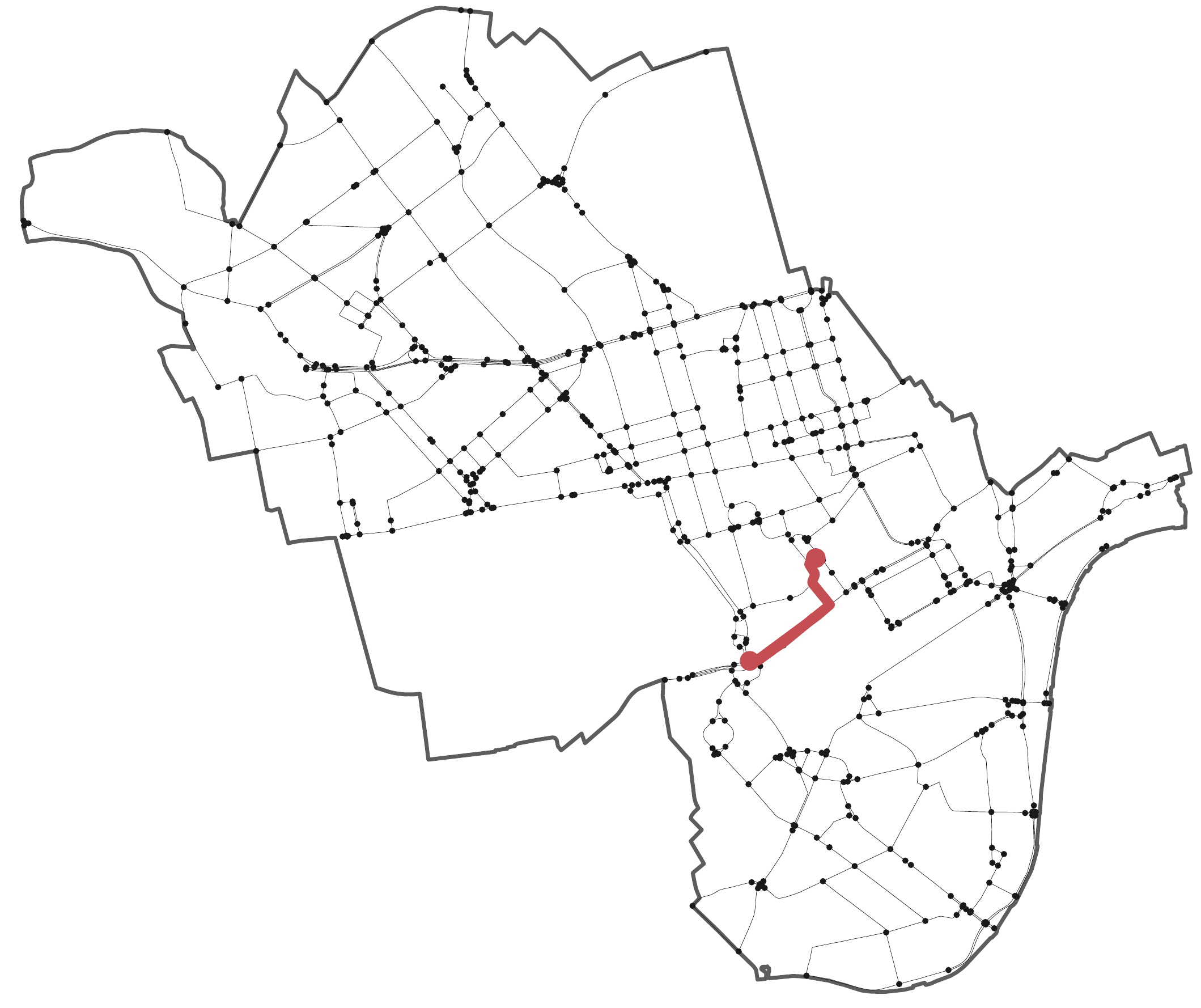}
    \vspace{-15pt}
    \caption{Example main road network with sample search path shown in red.}
    \label{fig:westminster}
    % \vspace{-6em}
\end{wrapfigure}

We divide the network into approximately equal length line-segments as discussed by \citet{shiode2020network} and shown in \cref{fig:westminster} (see also \cref{sec:spatial_locs}). The search regions are generated by defining a lower and upper bound of path length to search on the network. Instead of searching rectangular regions, we search over multiple, overlapping paths that lie on $\mathcal{G}$. The number of paths generated is very much dependent on the shape of the network and the lower and upper bounds for the path length. Although it is expected that the number of search paths overwhelming exceeds the number of rectangular search regions, the ability to easily hash network edges in dictionary-like data structures enables a simpler and faster implementation than its planar counterpart. For directional data sets (e.g. traffic sensors with road direction data), it is also possible to adapt the network-based scan to produce scores for each direction of a search path $S$. %Deducing the direction of travel in which activity is higher than expected, could be of great use to policy makers. %For non-directional data sets, we would usually assign four aggregated quantities, namely, $B_S, C_S, \aggbaselinelow, \aggbaselineupp$ to each search region. For directional data sets, this is easily modified by assigning eight quantities to each edge of the network; four for each direction. This effectively doubles the number of possible search regions but adds no extra computational complexity \todo{why?}. It is deemed that for some applications, the direction of travel may aid decision makers \todo{give an example of when it would}.
    
\subsection{Cluster likelihood metrics}
\label{sec:cluster_metrics}
We denote the collection of spatio-temporal search paths by $\mathcal{S}$. For every search region $S\in\mathcal{S}$, we define $B_S := \sum_{S}\baseline$ and $C_S := \sum_{S}\counts$ as the total baseline estimates and actual counts within a spatio-temporal region $S$ respectively. $\aggbaselineupp$ and $\aggbaselinelow$ are defined similarly. Fundamentally, for each searched space-time region $S$, we are interested in estimating the posterior probabilities $\mathbb{P}[H_1(S)\ |\ D]\label{prob_1}$ and $\mathbb{P}[H_0\ |\ D]\label{prob_2}$ for some null and alternative hypotheses $H_0$, $H_1(S)$ for $S\in\mathcal{S}$. The EBP model works under the null hypothesis of true counts being \emph{equal} to a Poisson distributed random variable with mean given by the baseline estimate. The alternative set of hypotheses are similar but generated with a mean of $q\baseline$ for some risk factor $q > 1$, which is to say:
\begin{align}
    H_0 &: \counts \sim \text{Po}(\baseline)\ \forall\ i = 1, \dots, N_s, \forall\ t\in\forecast\\
    H_1(S) &: \counts \sim \text{Po}(q \baseline)\ \forall\ c_i^t \in S \text{ and } \counts \sim \text{Po}(\baseline)\ \forall\ c_i^t \notin S \text{ for some } q > 1.\label{alt_hypoth}
\end{align}
Without using Bayesian approaches as in \citet{neill2010multivariate}, it is hard to calculate these probabilities in isolation.
Instead, \citet{neill2009expectation} introduces the Poisson likelihood metric, $\ebpscore(S)$, derived from their ratio (see \cref{sec:PLM} for a derivation). Note that in the case of $C_S < B_S$, the value of this metric defaults to 1. We introduce a generalised form of this metric which will also account for quieter activity than expected. We define the ASYM metric as
\begin{equation}
    \asymscore(S) := \begin{cases} 
                                \ebpscore(S) - 1, & 0 \leq B_S \leq C_S \\
                                1 - \ebpscore(S), & C_S \leq B_S
                           \end{cases}.
\end{equation}
The ability to specify different training profiles $\train$ motivates the introduction of this metric. Previous works have focused on capturing events with \emph{increased} counts; in the context of a COVID-19 monitoring system, this may not be the user's goal.
     
\section{Evaluation}
\label{sec:experiments}

\begin{wraptable}[7]{r}{0.48\linewidth}
    \centering
    \vspace{-20pt}
    \setlength{\tabcolsep}{4pt}
    \caption{Run-time metrics for each scan type.\label{tab: times}}
    \vspace{-6pt}
    \small
      \begin{tabular}{l l l l l }
      \toprule
        Type & Forecast time & $N_s$ & $|\mathcal{S}|$ & Scan time\\
        \midrule
        PL-GP  & $\sim$3h & \multirow{2}{*}{896} & \multirow{2}{*}{$32$k} & \multirow{2}{*}{$\sim$2min}\\
        PL-HW  & $\sim$15min &      &  & \\
        \midrule
        NET-GP & $\sim$1.5h & \multirow{2}{*}{$652$} & \multirow{2}{*}{$810$k} & \multirow{2}{*}{$\sim$4min}\\
        NET-HW & $\sim$\textbf{10min} &  & &\\
        \bottomrule
    \end{tabular}
\end{wraptable}

In the live version of the system, currently deployed in the city of London, we use real heterogeneous data sources such as loop detectors (for counting vehicles) and traffic cameras. That data is proprietary and hence cannot be included in this paper (a snippet of real traffic-camera data is shown in \cref{fig:jam_cam_forecast}, \cref{sec:appendix_forecasting}). That being said, to test the proposed framework, we use semi-synthetic, hourly vehicle count data from $\sim$900 road sensors across the borough of Westminster, London. See \cref{generating simulation data} for data generation process. We build the forecasts using 21 days worth of training data and set a forecasting period of $W=48$ hours. It is deemed that scanning over spatio-temporal regions which are 2 days long is the optimal length for detecting hotspots within this type of data.

\begin{figure}[ht!]
    % \vspace{-10pt}
    \centering
    \begin{subfigure}[t]{\textwidth}
        \centering
        \includegraphics[width=\textwidth]{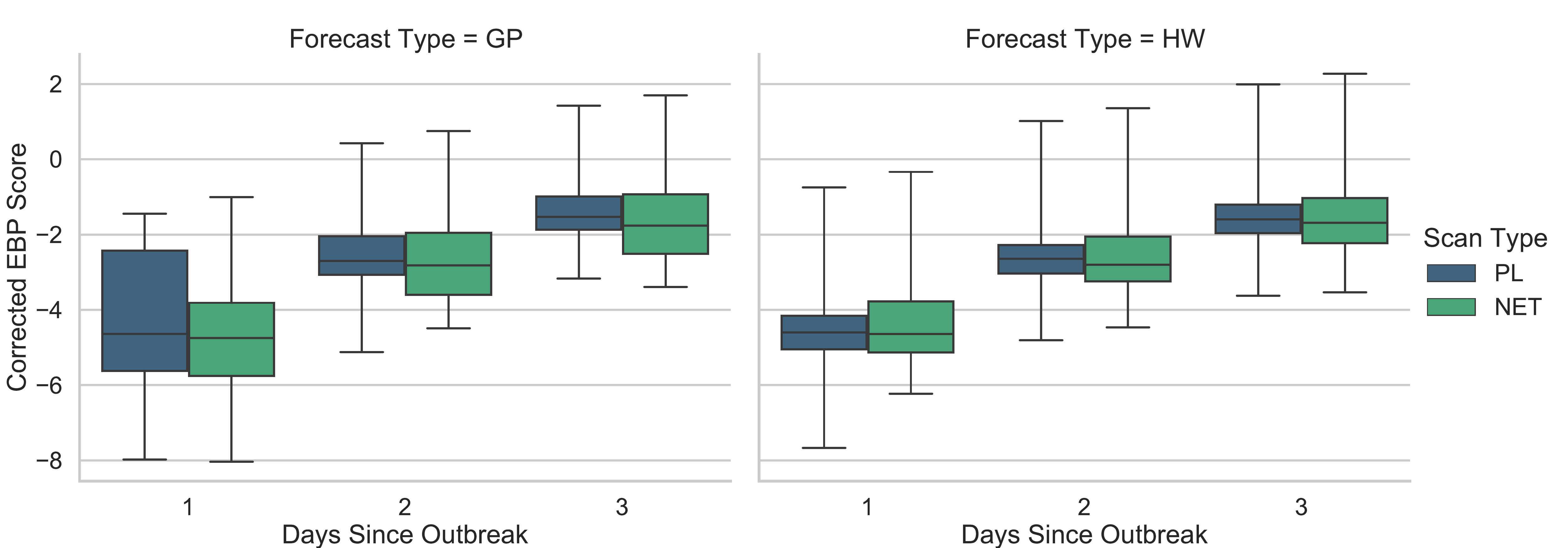}
        \caption{A higher corrected EBP score\footnotemark corresponds to a more significant result when compared to the 99th percentile of historical data (see \cref{generating simulation data}). We see here that on average, the NET scan yields scores of higher or similar significance for each day throughout the simulated surge.}
        \label{fig:days_to_detect}
    \end{subfigure}%
    \\
    \begin{subfigure}[t]{0.45\textwidth}
        \centering
        \includegraphics[width=\textwidth]{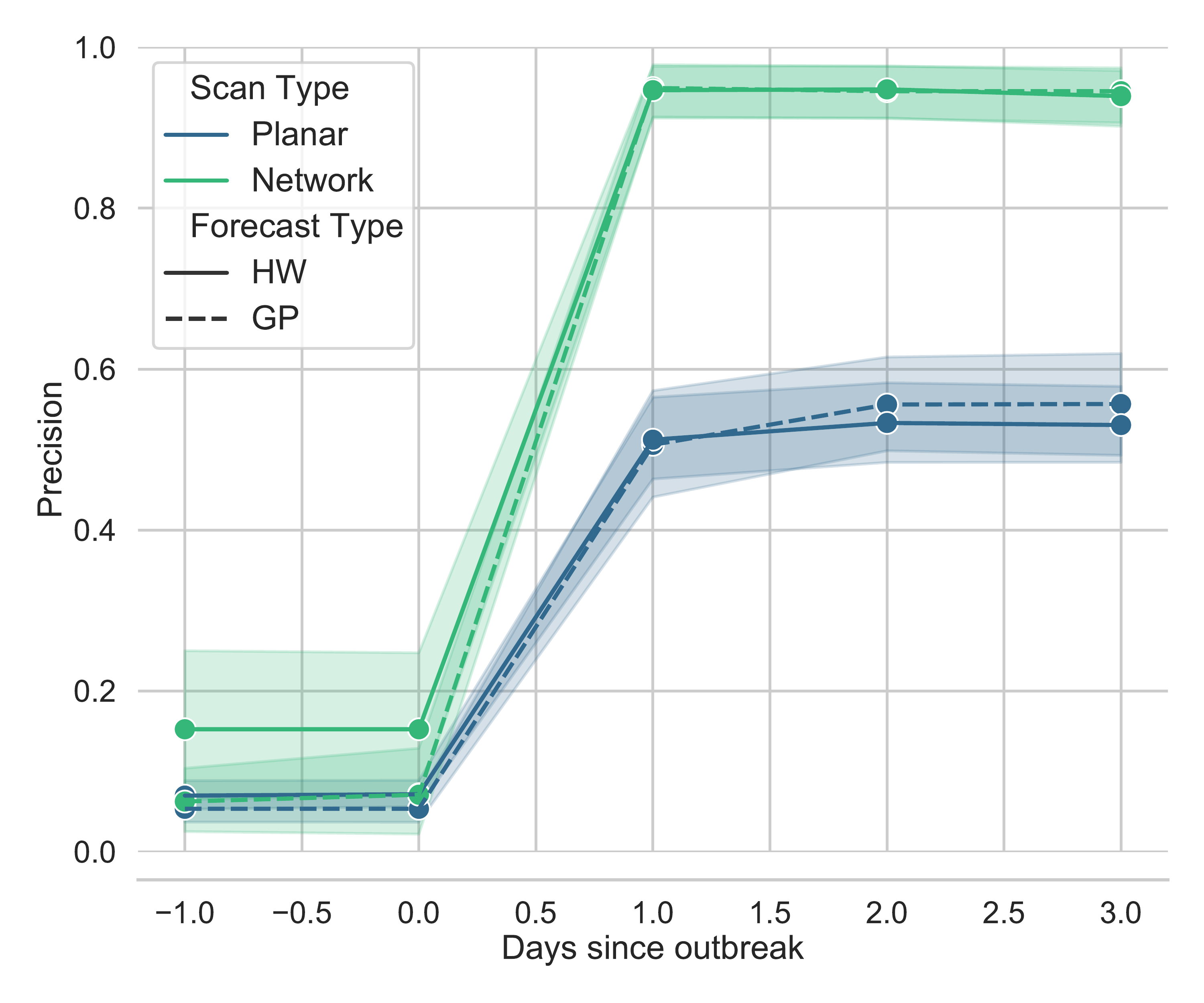}
        \caption{The NET scan demonstrates a higher spatial precision, i.e. the most significant search path returned from the scan contains a higher proportion of true surging detectors on average.}
        \label{precision}
    \end{subfigure}%
    \hspace{1em}
    \begin{subfigure}[t]{0.45\textwidth}
        \centering
        \includegraphics[width=\textwidth]{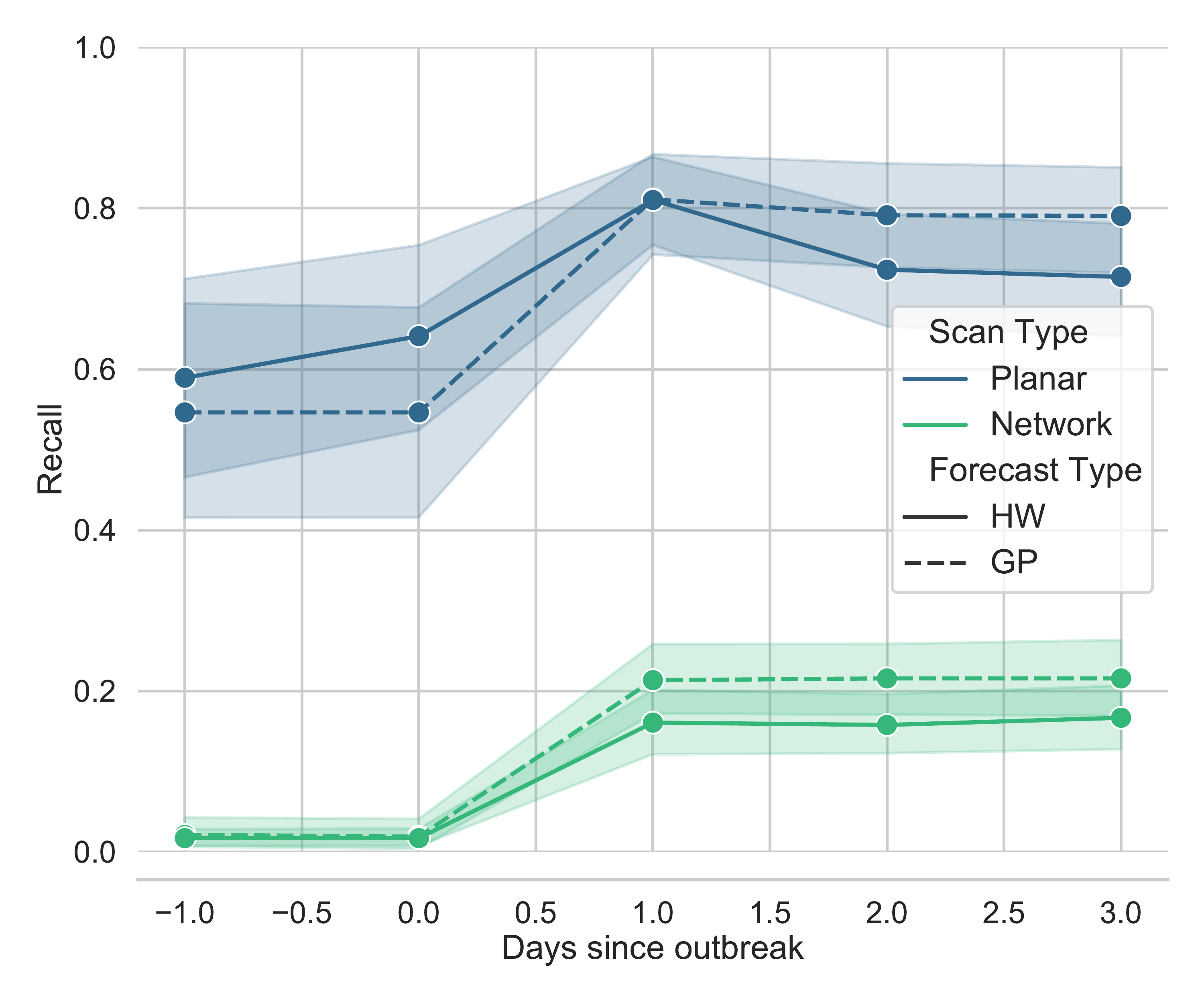}
        \caption{The NET scan suffers from a lower recall score due to the way that paths are constructed. i.e. the most significant search path returned from the scan is not representative of the true surging region.\label{recall}}
    \end{subfigure}
    \caption{A summary of scan statistic results for both HW/GP forecast types and PL/NET scans. \Cref{fig:days_to_detect} summarises the temporal performance, \Cref{precision} and \cref{recall} summarise the spatial performance.
    \label{fig:results}}
    \vspace{-2.9em}
\end{figure}

% \begin{figure}[ht!]
% \centering
% \includegraphics[width=0.8\textwidth]{figures/scan_comparison_single_alarm.pdf}
% \caption{NEEDS CHANGING: Scan results}
% \label{fig:GP scans results}
% \end{figure}

%   \begin{wrapfigure}{r}{0.5\textwidth}
%     \vspace{-10pt}
%     %\hspace{-20pt}
%     \label{something}
%     \centering
%     \includegraphics[width=0.5\textwidth]{figures/planar_heatmap.pdf}
%     \caption{DRAFT: Needs changing - Heatmaps}
%     \end{wrapfigure}
% \begin{wrapfigure}{r}{0.5\textwidth}
%     \centering
%     \vspace{-12pt}
%     \includegraphics[width=0.5\textwidth]{figures/mapboxlondon.png}
%     \caption{Scan Statistic Planar Results}
%     \label{fig:london}
% \end{wrapfigure}
    
For consistency, we define search regions that take roughly the same time to scan over for each scan type. For the network (NET) scan, we generate all paths on the main road network of Westminster that are between $50$m and $1$km in length; we include the lower bound to ensure that small circular paths on roundabouts are avoided. We also make the intermediate step of restricting the data set to detectors that are within $5\times 10^{-4}$ degrees of the main road network. For the planar (PL) scan, we first extend the Westminster boundary to its smallest bounding box and define a grid of resolution $N=8$. The number of detectors and search regions for each scan type are summarised in \cref{tab: times}.

\subsection{Results}
We compare the performance of each scan type by analysing three statistics; the average time to detect a simulated surge and its corresponding spatial \emph{precision} and \emph{recall} as defined in \cite{neill2009expectation}. \Cref{fig:days_to_detect} shows how the EBP score changes throughout the duration of the surge for each scan type. On average, the network scan yields scores of higher or similar significance when compared to its planar counterpart. However, it is likely to be reporting on a subset of the affected region due to the way paths are constructed. \Cref{precision} and \cref{recall} summarise the spatial accuracy of each scan type. In sum, if the required output of the system is a \emph{single} region of space-time that engulfs surging sensors, then it is deemed that the PL scan is superior in terms of speed, simplicity and high spatial recall. If the user is not interested in a singular return region and requires a more visual heat map output of the domain, then the NET scan offers a much more localised view of the surge; boasting superior spatial precision. For a more in-depth discussion on the results, the system and the future see \cref{sec:more_discussion}.

\footnotetext{A measure of how much the actual EBP score exceeds the alarm level (the 99th percentile).}

\section*{Acknowledgements and Disclosure of Funding}
Funded by Lloyd’s Register Foundation programme on Data Centric Engineering and Warwick Impact Fund via the EPSRC Impact Acceleration Account. Further supported by the Greater London Authority, Transport for London, Microsoft, Department of Engineering at University of Cambridge and the Science and Technology Facilities Council.

\bibliographystyle{icml2020}
\bibliography{bibliography}

\begin{thebibliography}{18}
\providecommand{\natexlab}[1]{#1}
\providecommand{\url}[1]{\texttt{#1}}
\expandafter\ifx\csname urlstyle\endcsname\relax
  \providecommand{\doi}[1]{doi: #1}\else
  \providecommand{\doi}{doi: \begingroup \urlstyle{rm}\Url}\fi

\bibitem[Baddeley et~al.(2015)Baddeley, Rubak, and Turner]{baddeley2015spatial}
Baddeley, A., Rubak, E., and Turner, R.
\newblock \emph{Spatial point patterns: methodology and applications with R}.
\newblock CRC press, 2015.

\bibitem[Baddeley et~al.(2020)Baddeley, Nair, Rakshit, McSwiggan, and
  Davies]{baddeley2020analysing}
Baddeley, A., Nair, G., Rakshit, S., McSwiggan, G., and Davies, T.~M.
\newblock Analysing point patterns on networks—a review.
\newblock \emph{Spatial Statistics}, pp.\  100435, 2020.

\bibitem[Baddeley et~al.(2000)Baddeley, M{\o}ller, and
  Waagepetersen]{baddeley2000non}
Baddeley, A.~J., M{\o}ller, J., and Waagepetersen, R.
\newblock Non-and semi-parametric estimation of interaction in inhomogeneous
  point patterns.
\newblock \emph{Statistica Neerlandica}, 54\penalty0 (3):\penalty0 329--350,
  2000.

\bibitem[Badr et~al.(2020)Badr, Du, Marshall, Dong, Squire, and
  Gardner]{badr2020association}
Badr, H.~S., Du, H., Marshall, M., Dong, E., Squire, M.~M., and Gardner, L.~M.
\newblock Association between mobility patterns and covid-19 transmission in
  the usa: a mathematical modelling study.
\newblock \emph{The Lancet Infectious Diseases}, 2020.

\bibitem[Chatfield(1978)]{chatfield1978holt}
Chatfield, C.
\newblock The holt-winters forecasting procedure.
\newblock \emph{Journal of the Royal Statistical Society: Series C (Applied
  Statistics)}, 27\penalty0 (3):\penalty0 264--279, 1978.

\bibitem[Daley \& Vere-Jones(2007)Daley and Vere-Jones]{daley2007introduction}
Daley, D.~J. and Vere-Jones, D.
\newblock \emph{An introduction to the theory of point processes: volume II:
  general theory and structure}.
\newblock Springer Science \& Business Media, 2007.

\bibitem[Diggle(2013)]{diggle2013statistical}
Diggle, P.~J.
\newblock \emph{Statistical analysis of spatial and spatio-temporal point
  patterns}.
\newblock CRC press, 2013.

\bibitem[Gelfand et~al.(2010)Gelfand, Diggle, Guttorp, and
  Fuentes]{gelfand2010handbook}
Gelfand, A.~E., Diggle, P., Guttorp, P., and Fuentes, M.
\newblock \emph{Handbook of spatial statistics}.
\newblock CRC press, 2010.

\bibitem[Kulldorff(1997)]{kulldorff1997spatial}
Kulldorff, M.
\newblock A spatial scan statistic.
\newblock \emph{Communications in Statistics-Theory and methods}, 26\penalty0
  (6):\penalty0 1481--1496, 1997.

\bibitem[Matrajt \& Leung(2020)Matrajt and Leung]{matrajt2020evaluating}
Matrajt, L. and Leung, T.
\newblock Evaluating the effectiveness of social distancing interventions to
  delay or flatten the epidemic curve of coronavirus disease.
\newblock \emph{Emerging infectious diseases}, 26\penalty0 (8):\penalty0 1740,
  2020.

\bibitem[Muhammad et~al.(2020)Muhammad, Long, and Salman]{muhammad2020covid}
Muhammad, S., Long, X., and Salman, M.
\newblock Covid-19 pandemic and environmental pollution: a blessing in
  disguise?
\newblock \emph{Science of The Total Environment}, pp.\  138820, 2020.

\bibitem[Neill(2009)]{neill2009expectation}
Neill, D.~B.
\newblock Expectation-based scan statistics for monitoring spatial time series
  data.
\newblock \emph{International Journal of Forecasting}, 25\penalty0
  (3):\penalty0 498--517, 2009.

\bibitem[Neill(2019)]{neill2019machine}
Neill, D.~B.
\newblock \emph{Machine Learning and Event Detection for Population Health}.
\newblock Georgia Institute of Technology, 2019.

\bibitem[Neill \& Cooper(2010)Neill and Cooper]{neill2010multivariate}
Neill, D.~B. and Cooper, G.~F.
\newblock A multivariate bayesian scan statistic for early event detection and
  characterization.
\newblock \emph{Machine learning}, 79\penalty0 (3):\penalty0 261--282, 2010.

\bibitem[Neill et~al.(2005)Neill, Moore, Sabhnani, and
  Daniel]{neill2005detection}
Neill, D.~B., Moore, A.~W., Sabhnani, M., and Daniel, K.
\newblock Detection of emerging space-time clusters.
\newblock In \emph{Proceedings of the eleventh ACM SIGKDD international
  conference on Knowledge discovery in data mining}, pp.\  218--227, 2005.

\bibitem[Rasmussen \& Williams(2005)Rasmussen and Williams]{GPBook}
Rasmussen, C.~E. and Williams, C. K.~I.
\newblock \emph{Gaussian Processes for Machine Learning (Adaptive Computation
  and Machine Learning)}.
\newblock The MIT Press, 2005.
\newblock ISBN 026218253X.

\bibitem[Shiode \& Shiode(2020)Shiode and Shiode]{shiode2020network}
Shiode, S. and Shiode, N.
\newblock A network-based scan statistic for detecting the exact location and
  extent of hotspots along urban streets.
\newblock \emph{Computers, Environment and Urban Systems}, 83:\penalty0 101500,
  2020.

\bibitem[Wesolowski et~al.(2012)Wesolowski, Eagle, Tatem, Smith, Noor, Snow,
  and Buckee]{wesolowski2012quantifying}
Wesolowski, A., Eagle, N., Tatem, A.~J., Smith, D.~L., Noor, A.~M., Snow,
  R.~W., and Buckee, C.~O.
\newblock Quantifying the impact of human mobility on malaria.
\newblock \emph{Science}, 338\penalty0 (6104):\penalty0 267--270, 2012.

\end{thebibliography}

\newpage

\appendix

\section{The value and ethics of using `big-data' to capture human mobility}
\label{sec:more_discussion}

\begin{figure}[ht]
    \centering
    \includegraphics[width=\textwidth]{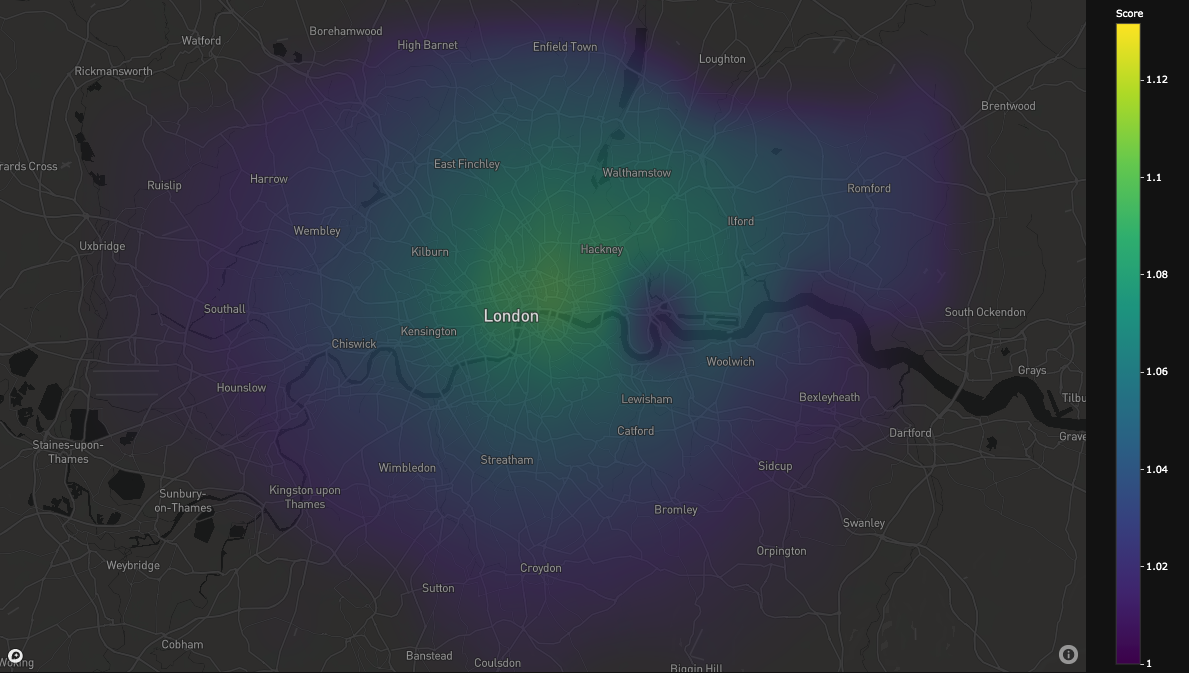}
    \caption{Early warning system preview. Planar scan statistic heatmap over London, UK. Regions of likely clusters are shown in yellow, using the simulated data.}
    \label{London Mapbox}
\end{figure}

\paragraph{Concluding remarks on the early warning system } It is clear that there are benefits and drawbacks to all types of scans which are mostly dependent on the scientific use case. Furthermore, the size of the region of interest also plays its part. The simplicity of the PL scan allows for much larger domains without increasing the run-time of the scan; for example, the whole city of London, UK. It was found that imposing a $16\times16$ grid on the entire London domain provided a sufficient resolution for detecting larger regions of increased activity -- shown in \cref{London Mapbox}. However, it is deemed that if data sources are confined to a network that sits in a relatively small region of space (e.g. a borough), then the NET scan is far superior in visualising localised hotspots with the added advantage of including the direction of the surged activity.

On the forecasting, the two forecasting methods, appear to perform similarly, again both with their benefits and drawbacks. The HW method is a good and fast starting point for creating these type of baselines, boasting a factor of 20 faster run time than a GP. However, if run-time is not an issue, GP's manage to achieve a slightly better spatial accuracy with the added benefit of uncertainty quantification. 

\begin{figure}[ht!]
    \centering
     \begin{subfigure}[t]{0.49\textwidth}
        \centering
        \includegraphics[width=\textwidth]{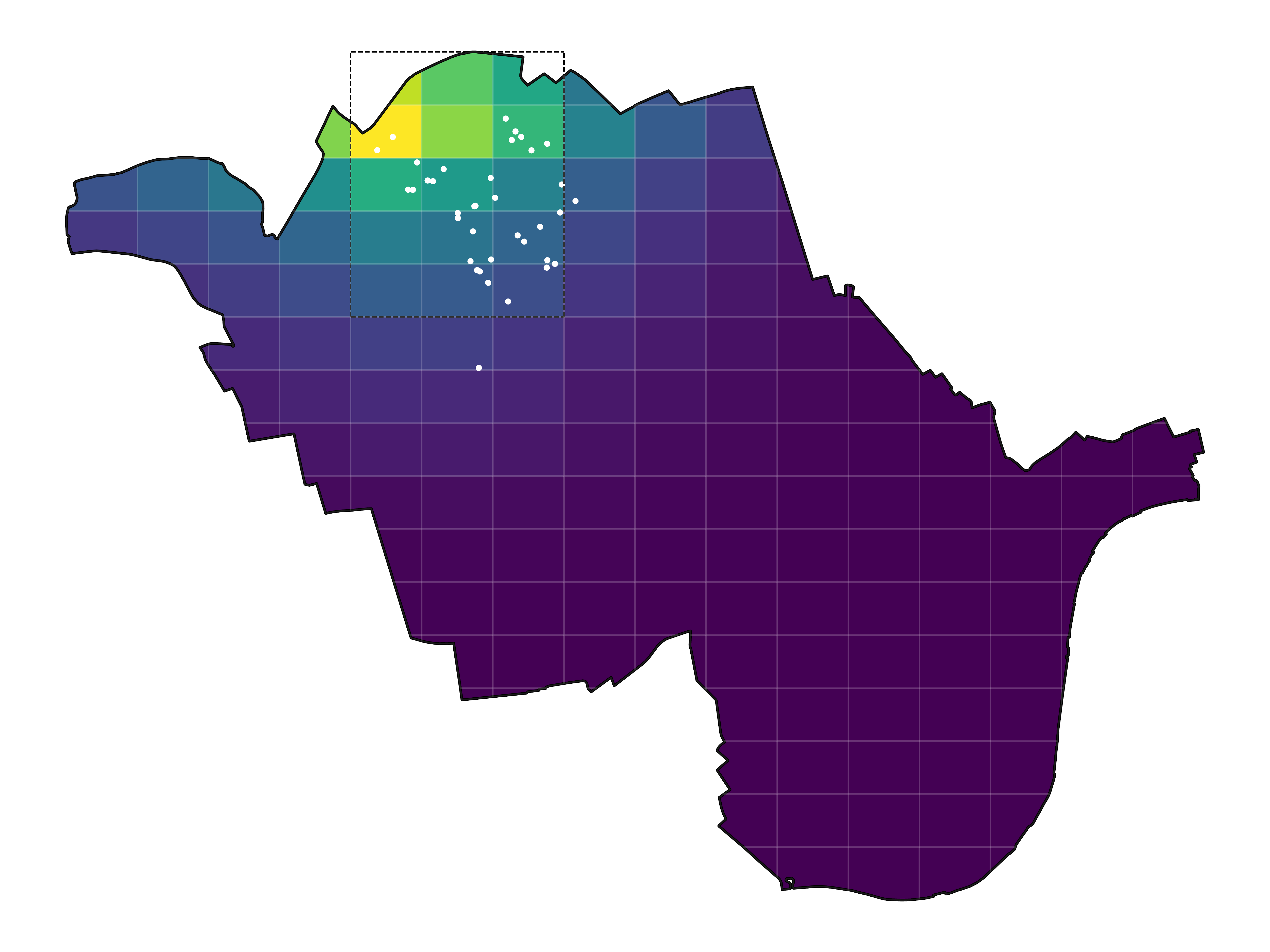}
        \caption{Example planar scan of all of Westminster borough in London. With 20 sensors picked for demonstration, in white.}
    \end{subfigure}%
    \hfill
    \begin{subfigure}[t]{0.49\textwidth}
        \centering
        \includegraphics[width=\textwidth]{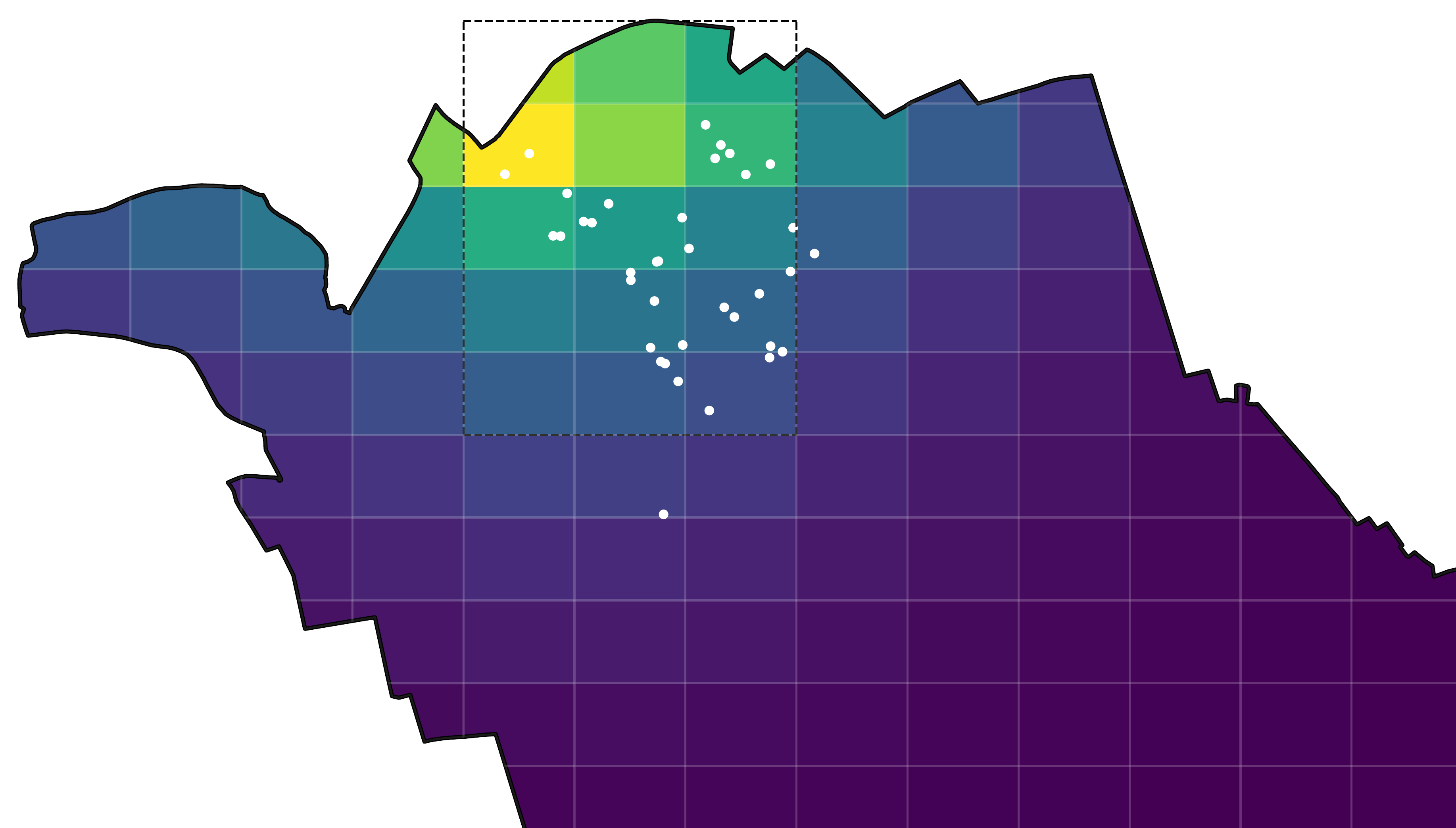}
        \caption{Example results from a planar scan affecting roughly 20 sensors. Surging sensor locations are shown in white and the spatial slice of the most significant search region $S$ in dashed black.}
        \label{planar scan heatmap}
    \end{subfigure}%
    \hfill
    \begin{subfigure}[t]{0.49\textwidth}
        \centering
        \includegraphics[width=\textwidth]{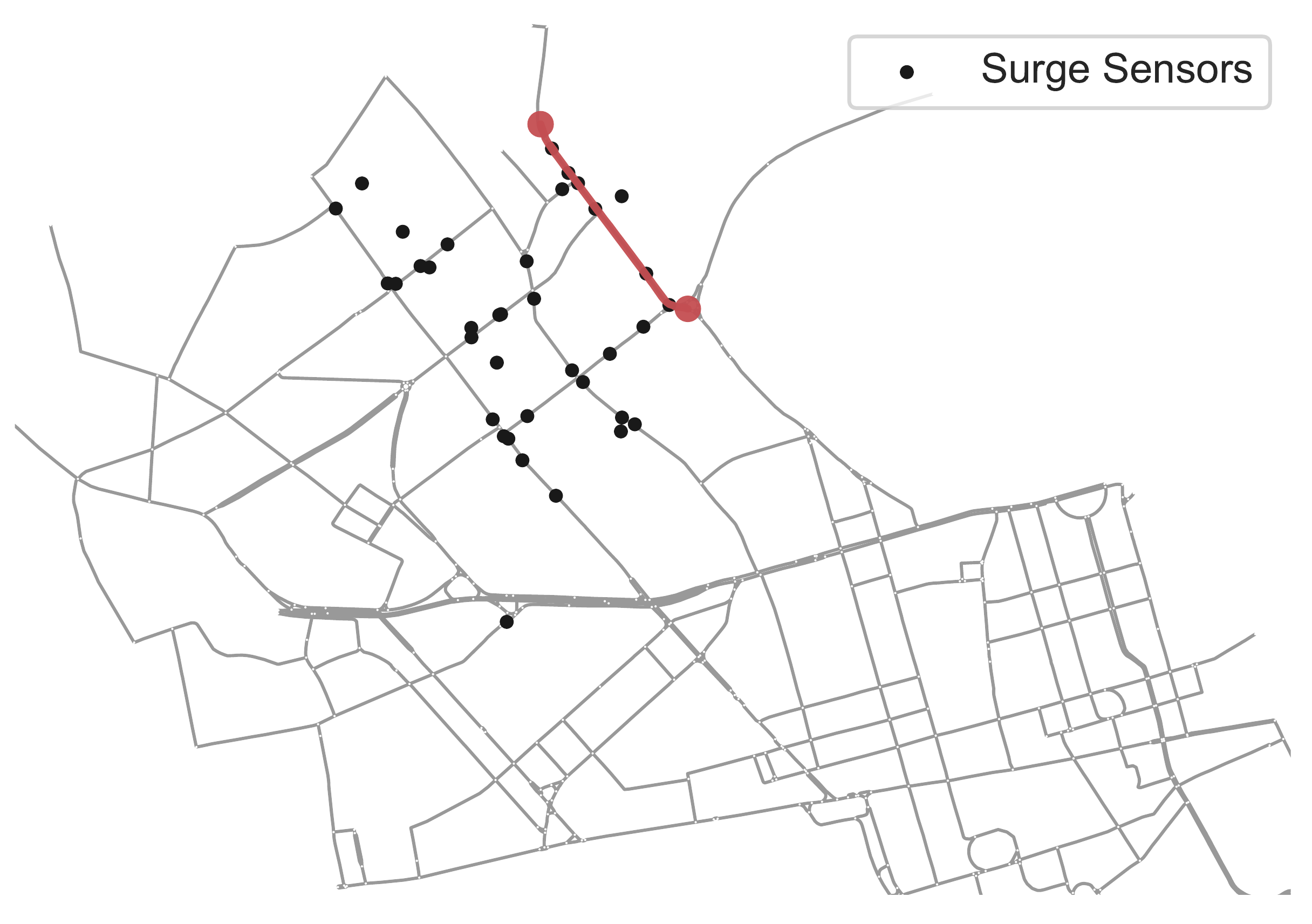}
        \caption{Road network sensors shown, on the same region as the planar scan in \cref{planar scan heatmap}. Included is also an example line-segment used by the NET scan. The full road network in Westminster borough is shown in \cref{fig:westminster}.}
    \end{subfigure}
    \hfill
    \begin{subfigure}[t]{0.49\textwidth}
        \centering
        \includegraphics[width=\textwidth]{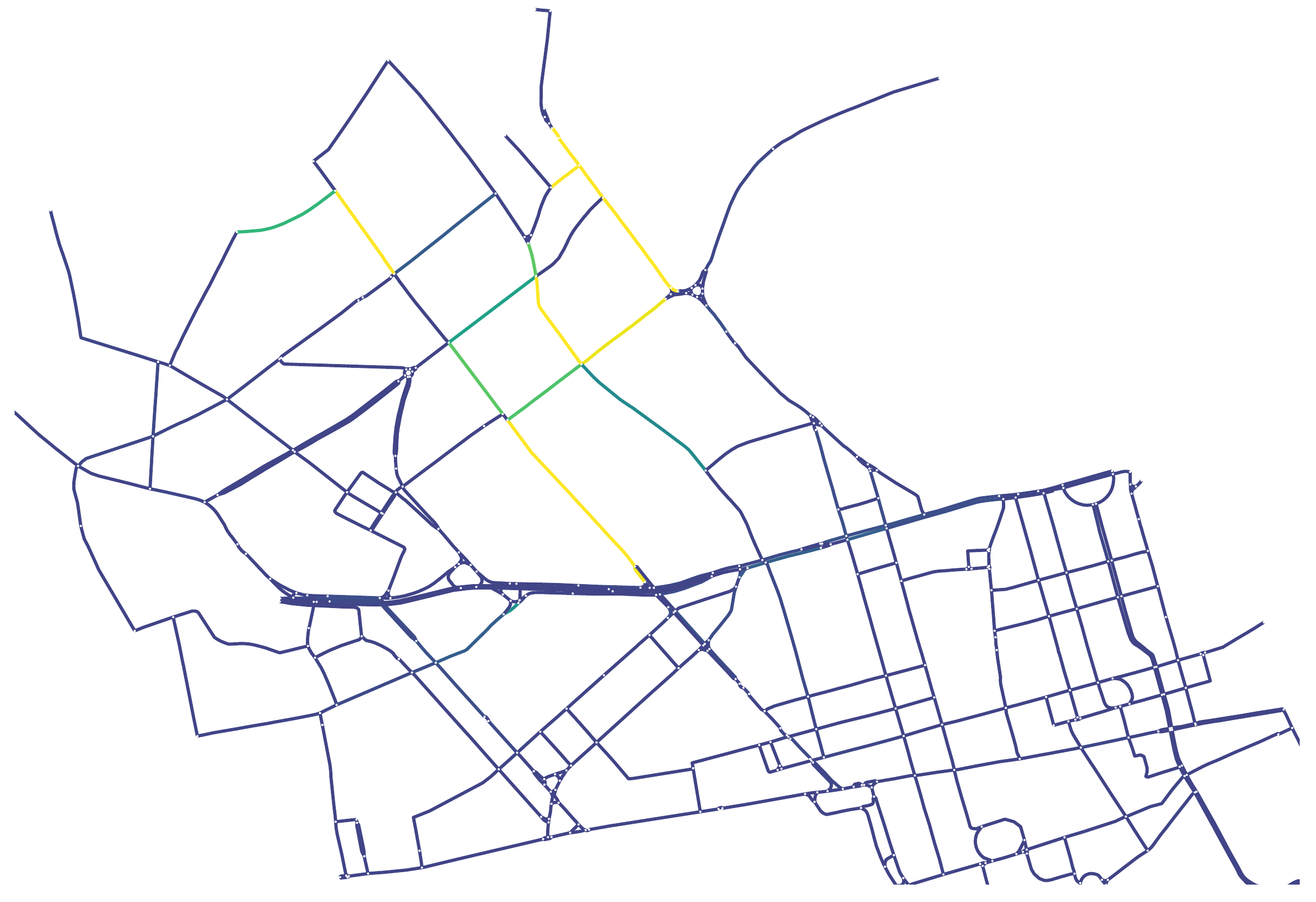}
        \caption{Network heatmap results for the same surging sensors as shown in \Cref{planar scan heatmap}. Regions of high significance (yellow) are clearly more localised than the planar counterpart.\label{fig: netscan sim result}}
    \end{subfigure}
    \caption{Example simulated surge results viewed in heatmap form}
    \label{fig:example_heatmap}
\end{figure}

Whilst both variants of the NET scan boast a higher spatial precision, the suffering recall metric suggests that a possibly more useful output is shown in a heat map form as shown in \cref{fig:example_heatmap}. These heat maps are generated by taking the collection of searched regions and aggregating by mean to the defined grid. This output should be of greater use to decision makers; providing a simple and visual explanation of the underlying dynamics. It also yields information about sensors which are \emph{not} surging which is something that is not available from a single returned search region. Naturally, presenting these heat maps with a temporal dependence (e.g. through the use of a slider) is even more useful in showing the evolution of a potential surge.

\FloatBarrier

\paragraph{London's digital twin of busyness} The early warning system presented herein, plays but a small part in the larger digital twin. The twin combines scalable cloud computing and machine learning to continuously learn and update itself by drawing from multiple streams of transportation data, capturing the constantly evolving activity in the city. A digital twin is more than a digital abstraction reflecting reality, rather it also attempts to generate interventions which may be directly implemented or via “human-in-the-loop” recommendations to advisors to benefit both accuracy of the abstraction and any objectives therein. One such example within London has been observing air quality within areas of high public transit. Encoding a capacity to represent pollution nearby a school or park therefore permits estimating alternative routes or prioritising hybrid buses for a given route. Other parts of the twin can count the number of people walking on a street and estimate their social distance from snapshots of Transport for London’s traffic camera footage.  This is implemented through specialised cloud compute clusters, virtual machines and data structures activity engaged in data collection, event detection, and intervention evaluation. A future extension of our work is to apply our scan statistics to multiple modes of transport thus giving further insights to policy makers.

\paragraph{Privacy} As was touched upon in \cref{sec:intro} privacy is a major issue when conducting large-scale human monitoring of the form done in this project. First, owing to the nature of a pandemic of this magnitude, spread and severity, certain civil liberties may need to be temporarily suspended or bypassed. Around the world this has been the case, where e.g. some lock-downs have been more onerous than others. Similarly for privacy, in order to get ahead of the virus transmission, certain liberties may need to be temporarily suspended in order for the scientific community to fully understand the virus. That being said, any new government powers must expire when the disease is contained and any new processing of personal data must be proportionate to the actual need. In this project all data is anonymised, collected data cannot be reverse engineered in order to track people and there is a strong regiment of ethical oversight. Ultimately though, we are not interested in micro-behaviours, but rather the macro behaviour of people as this is what is required for more informed policy decisions. 

Further, we have taken steps to make our work transparent and reproducible. Transparency improves trust in a system from the general public – particularly important when many are rightly concerned about user privacy and the ``big brother'' effect. By making our code open-source, using technologies such as Docker, and (where possible) the data publicly accessible, other researchers can reproduce our work and our methods are fully transparent.

\paragraph{Future work} On submission the digital twin, with its early warning system, is live in the city of London. Discussions are ongoing with regards to expanding the system to other cities in the UK. On the technical level there are a multitude of improvements that can be made as well as far more research undertaken. For the warning system alone, a more robust way of modelling the busyness would be to employ an inhomogeneous spatio-temporal point-process model, which would allow for non-stationary intensity function -- see e.g. the work by \citet{baddeley2000non}. That being said, elegant though that approach may be, the complexity of such a model is orders of magnitude higher than our initial proposal herein. Moreover, the technical improvements must be weighed against the ultimate users of this system i.e. the stakeholders, for whom usability if the primary motivator. Under this light, a research software engineering direction is actively focused on ease of use, progressing Application Programming Interface (API) simplicity and model transparency. 

\paragraph{Stakeholders} The two primary stakeholders are the Greater London Authority and Transport for London. When explaining our work to policy makers we emphasise high level mechanics, intuition and limitations of methods rather than the mathematical intricacies. As the system matures, it is our hope that the list of stakeholders grows, so that multiple policy makers can make use of the system. We see this as a likely future scenario as the virus, unfortunately, appears to be the `new normal'.

\clearpage
\section{Spatial locations on a grid}
\label{sec:spatial_locs}
\begin{figure}[ht]
    \centering
    \includegraphics[width=\textwidth]{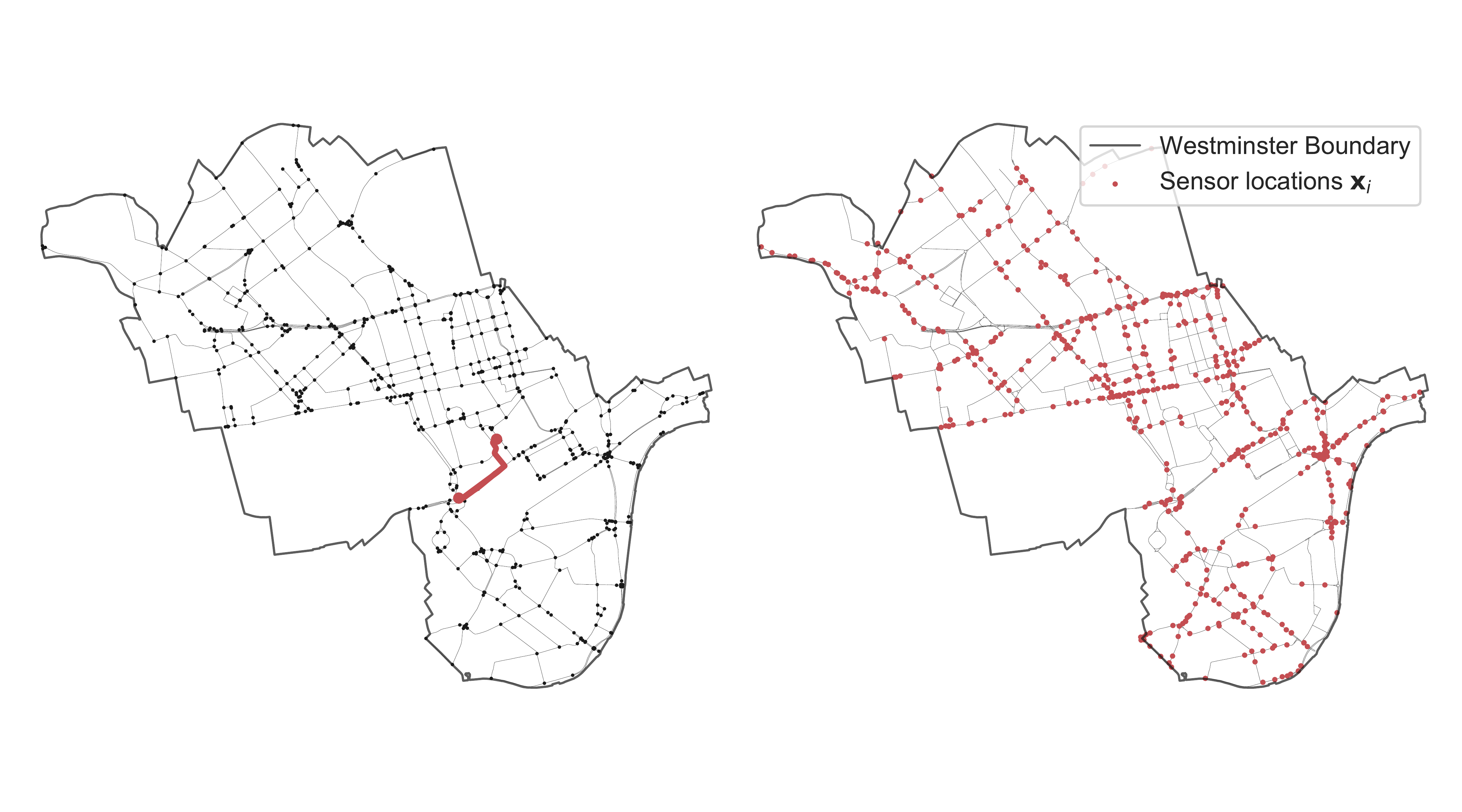}
    \caption{Surge sensor on road network. The \textbf{left} panel shows the road network in Westminster borough in London, UK. Also shown is an example path used for the NET scan statistics model. The \textbf{right} panel shows the same borough but this time the surge sensors are superimposed on the road network, where it can be seen that larger thoroughfare have more surge sensors (which is to be expected).}
\end{figure}
\FloatBarrier

\clearpage
\section{Time-series forecasting}
\label{sec:appendix_forecasting}
\begin{figure}[ht]
    \centering
    \includegraphics[width=\textwidth]{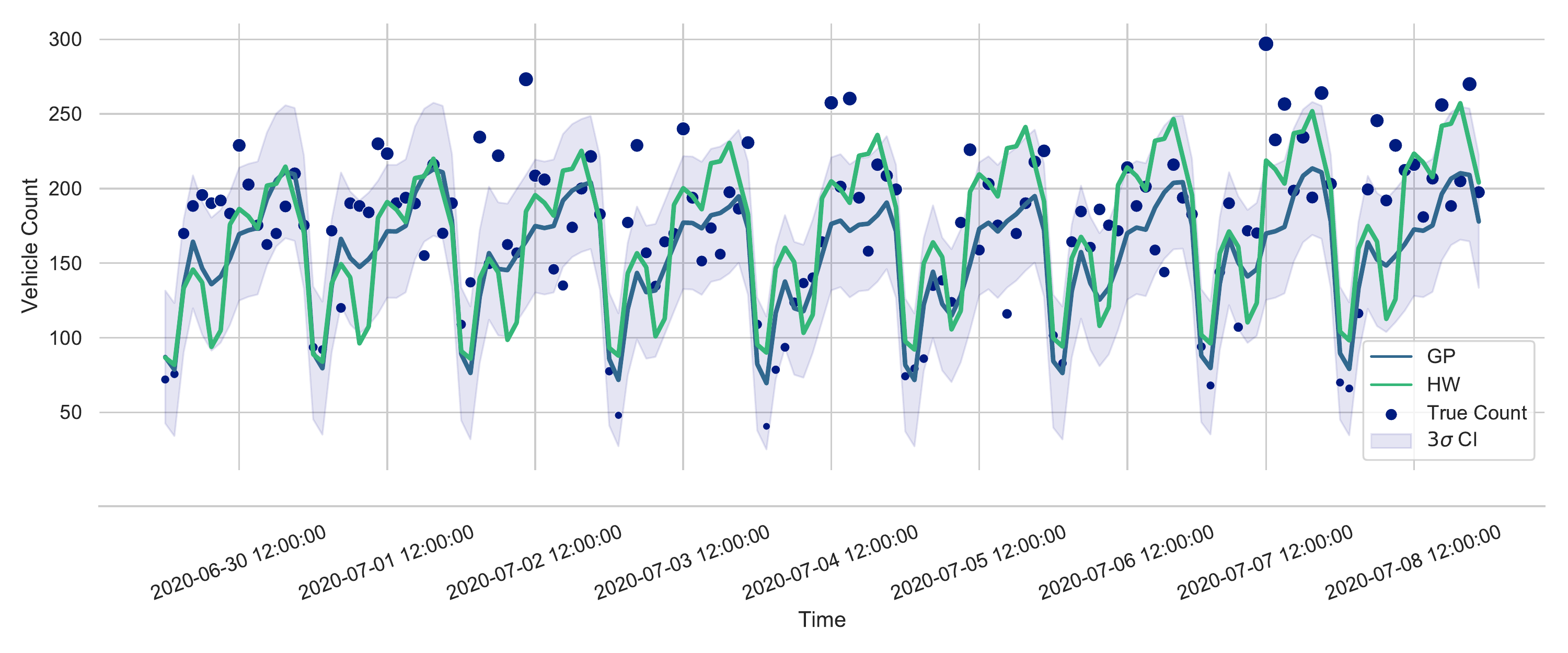}
    \caption{A comparison of `busyness' forecasting methods used within the scan statistic pipeline, when applied to data from London's JamCam system -- which captures how traffic is moving along particular roads and helps people to understand live traffic conditions. Both methods appear to perform well with clear periodic variation in both. In this particular example, the HW method appears to exaggerate some of this periodic behaviour more severely than the GP counterpart. Clearly, the GP method comes with the added benefit of confidence bands. Throughout what follows, we use a 3$\sigma$ confidence interval.}
    \label{fig:jam_cam_forecast}
\end{figure}

The Holt-Winters (HW) method \cite{chatfield1978holt} is trained on historical counts $\{\counts \ | \ t \in \train\}$ by iterating the three equations given by \cref{eq:HoltWinters}. It has three main components: the smoothed value $X_t$, trend component $Y_t$, and the periodic component $Z_t$
\begin{equation}
\label{eq:HoltWinters}
\begin{array}{l}
X_{t}=\alpha \frac{c_{i}^{t}}{Z_{t-24}}+(1-\alpha)\left(X_{t-1}+Y_{t-1}\right) \\
Y_{t}=\beta\left(X_{t}-X_{t-1}\right)+(1-\beta) Y_{t-1} \\
Z_{t}=\gamma \frac{c_{i}^{t}}{X_{t}}+(1-\gamma) Z_{t-24} 
\end{array}
\end{equation}
where
\begin{equation}
\label{eq:FinalHoltWinters}
b_i^t=\left(X_{t-1}+Y_{t-1}\right) Z_{t-24}.
\end{equation}
Tuning parameters $\alpha$, $\beta$, and $\gamma$ are optimised to increase the forecast accuracy of the model. The baseline estimate for the next hour is then given by \cref{eq:FinalHoltWinters} above. The whole training period is used to predict $\baseline$ for the next hour, and hence, by iterating this process, the baseline for any subsequent hour can be determined. An example of HW applied to real count data is shown in \cref{fig:jam_cam_forecast}.

\clearpage
\section{The Poisson likelihood metric}
\label{sec:PLM}
The Poisson likelihood metric, $\ebpscore(S)$ is derived from the ratio of $\mathbb{P}[D| H_0]$ and $\mathbb{P}[D| H_1 (S)], S\in\mathcal{S}$. In its original form, the metric is derived as follows:
\begin{align}
        \ebpscore(S) &:= \dfrac{\mathbb{P}[D| H_1 (S)]}{\mathbb{P}[D| H_0]}\\
             & = \dfrac{\max_{q > 1}\prod_{\counts \in S}\mathbb{P}(\counts \sim \text{Po}(q\baseline))}{\prod_{\counts \in S}\mathbb{P}(\counts \sim \text{Po}(\baseline))}\\
             & = \max_{q > 1}\prod_{\counts \in S}e^{(1-q)\baseline}q^{\counts}\\
             & = \max_{q > 1}e^{(1-q)B_S}q^{C_S}\\
             & = \bigg(\dfrac{C_S}{B_S}\bigg)^{C_S} e^{B_S - C_S}
\end{align}
using that the maximum value of $\ebpscore(S)$ is achieved for $q = \max(1, C_S / B_S)$.

\clearpage
\section{Generating Simulation Data}
\label{generating simulation data}

In this example, we aim to create data that corresponds to a \emph{mid-lockdown} profile beginning from April 2020 and focus on identifying regions of \emph{increased activity}. %Indeed, this example only includes analysis of the original EBP metric.% We focus on comparing the two main forecasting methods discussed in \cref{forecasting} and for both instances,
To generate (and anonymise) the surge-free data, we first calculate the 90th percentile of the amplitude for each sensor and use this to generate a sinusoidal time-series with added Poisson noise. The periods of the sinusoidal variation are chosen to match the empirical daily and weekly variation of each sensor. We use this generated data for two purposes; firstly to act as base data to later build a surge on top of and secondly, to build up the empirical distribution of the EBP metric when the scan is performed on surge-free data. In practise, real data requires the need for pre-processing; our implementation includes a four-stage pipeline for this which involves anomaly detection/removal, missing data interpolation and the removal of time-series which are deemed to be either a-periodic or missing too much data.

To simulate a surge, we first choose an epicentre uniformly at random within the borough boundary and it's $k$ nearest sensors; $k$ is chose uniformly at random between $k_{\text{min}}=10$ and $k_{\text{max}} = 100$. We apply a linear increase in each of these affected sensor's time-series spanning the last three days; it is this three day surge that we wish to capture in the scan. We give each sensor its own rate of linear increase $\omega_i$ which is proportional to it's busyness i.e. busier sensors will surge faster than quieter ones. This was found to give a similar surge severity to what would be expected from a breach in lockdown rules. Parametrically, the surge factor $\lambda_i(t) = (1 + \omega_it)$ for outbreak days $t=1, 2,3$ is bounded between 1 and 4.

\begin{figure}[hb]
    \centering
    \includegraphics[width=0.9\textwidth]{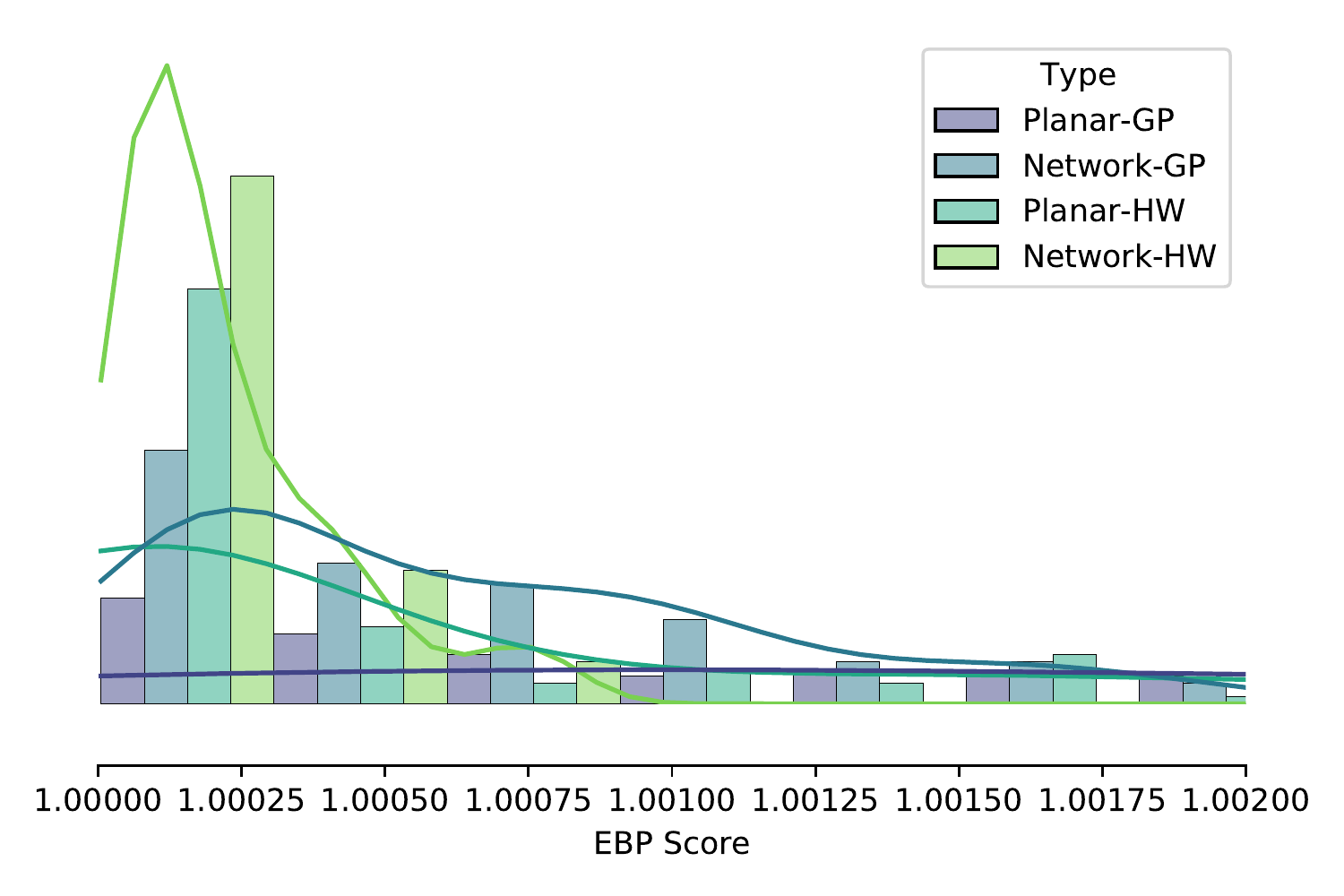}
    \caption{Distribution of EBP scores from surge free simulated data. Both variations of the NET scan appear yield a thinner tailed distribution, potentially making ``significant" easier to define in the alarm system.}
    \label{fig: surge free distributions}
\end{figure}

To define when a surge is \emph{detected}, we first estimate the empirical distribution of EBP scores when no surges are present. Using the mid-lockdown profile, we generate 122 days worth of surge-free sensor data across the borough; the first 21 days are used to train the model for the first forecast period. Using the remaining 101 days data, we calculate the highest EBP score for each scan type; the distributions are shown in \Cref{fig: surge free distributions}. It is this empirical distribution that is used to calculate the scan's average time to detection as a function of allowable false-positive rate.

\end{document}